\documentclass[conference]{IEEEtran}
\IEEEoverridecommandlockouts

\usepackage[noadjust]{cite}

\usepackage{amsmath,amssymb,amsfonts}
\usepackage{algorithmic}
\usepackage{graphicx}
\usepackage{textcomp}
\usepackage{xcolor}
\usepackage{microtype}
\usepackage{hyperref}
\usepackage{eso-pic}
\def\BibTeX{{\rm B\kern-.05em{\sc i\kern-.025em b}\kern-.08em
    T\kern-.1667em\lower.7ex\hbox{E}\kern-.125emX}}

\AddToShipoutPictureBG*{%
  \AtPageUpperLeft{%
    \hspace{2cm}%
    \raisebox{-14pt}{%
      \makebox[0pt][l]{%
        \begin{minipage}{13cm}
          \fontsize{7}{8}\selectfont This paper has been accepted for publication in Proc. IEEE ICTAI 2025, Athens, Greece. This is the author's version which has not been fully edited and content may change prior to final publication. Citation information: DOI 10.1109/ICTAI66417.2025.00064
        \end{minipage}
      }
    }
  }
  \AtPageLowerLeft{%
    \hspace{5cm}%
    \raisebox{32pt}{%
      \makebox[0pt][l]{%
        \begin{minipage}{10cm}
          \fontsize{7}{8}\selectfont ©2025 IEEE. Personal use of this material is permitted. Permission from IEEE must be obtained for all other uses, in any current or future media, including reprinting/republishing this material for advertising or promotional purposes, creating new collective works, for resale or redistribution to servers or lists, or reuse of any copyrighted component of this work in other works.
        \end{minipage}
      }
    }
  }
}

\begin{document}

\title{Feature Space Topology Control via Hopkins Loss}

\author{\IEEEauthorblockN{Einari Vaaras}
\IEEEauthorblockA{\textit{Signal Processing Research Centre} \\
\textit{Tampere University}\\
Tampere, Finland \\
einari.vaaras@tuni.fi}
\and
\IEEEauthorblockN{Manu Airaksinen}
\IEEEauthorblockA{\textit{BABA Center, Department of Physiology} \\
\textit{University of Helsinki}\\
Helsinki, Finland \\
manu.airaksinen@helsinki.fi}
}

\maketitle

\begin{abstract}
Feature space topology refers to the organization of samples within the feature space. Modifying this topology can be beneficial in machine learning applications, including dimensionality reduction, generative modeling, transfer learning, and robustness to adversarial attacks. This paper introduces a novel loss function, Hopkins loss, which leverages the Hopkins statistic to enforce a desired feature space topology, which is in contrast to existing topology-related methods that aim to preserve input feature topology. We evaluate the effectiveness of Hopkins loss on speech, text, and image data in two scenarios: classification and dimensionality reduction using nonlinear bottleneck autoencoders. Our experiments show that integrating Hopkins loss into classification or dimensionality reduction has only a small impact on classification performance while providing the benefit of modifying feature topology. The code is freely available at \url{https://github.com/SPEECHCOG/hopkins_loss}.
\end{abstract}


\section{Introduction} \label{sec_introduction}

The concept of \textit{feature space topology} refers to how samples are organized in the feature space. In the context of machine learning (ML), modifying features into a desired topology can be potentially useful in multiple contexts: First, imposing a specific structure on the feature space during ML model training might improve generalization. For example, regularly-spaced features might help in reducing overfitting by ensuring that the features are uniformly distributed \cite{rethinking_feature_distribution_image_classification, topoloss}. Second, when using nonlinear bottleneck autoencoders (AEs), transforming high-dimensional data into a lower-dimensional feature space with a desired feature topology may be beneficial for e.g. visualization, data compression, or as a pre-processing step for other ML algorithms such as for classification purposes \cite{dim_reduction_invariant_mapping, improving_dim_reduction_projections, dim_reduction_task, automated_data_mining_framework_autoencoder_dim_reduction}.

Third, controlling the distribution of generated features in generative models, such as generative adversarial networks \cite{gan_original}, may be useful in applications such as image synthesis, where the spatial arrangement of features is important \cite{controllable_gans, focal_freq_loss_image_synthesis, gans_learn_distributions}. Fourth, constructing features with specific topological properties may benefit domains like bioinformatics, where the spatial arrangement of features (e.g. genes) can be critical \cite{feature_engineering_protein_functions, taxonomy_aware_feature_engineering}. Fifth, in transfer learning, adjusting the feature space topology of the source or target domain can help in aligning the respective feature spaces \cite{transfer_learning_feature_space_mapping, transfer_learning_algorithm_evaluation_ieee_ictai, domain_adaptation_feature_whitening, heterogeneous_transfer_learning}. Finally, controlling feature space topology might make ML models more robust to adversarial attacks, as the topology of the feature space may affect the sensitivity of ML models to perturbations \cite{topology_layer, improving_adversarial_robustness_feature_spectral_regularization, graph_neural_networks_topological_perturbations}.

In this paper, we propose a novel loss function called Hopkins loss, which is based on the Hopkins statistic ($H$, see Section \ref{subsec_hopkins_statistic}). This loss function can be integrated into ML model training to modify or enforce the feature space into either a regularly-spaced, randomly-spaced, or clustered topology. This is in contrast to existing topology-related methods (Section \ref{subsec_feature_space_topology_methods}) that all aim to preserve the topology of input features rather than try to transform it into a user-defined form. We evaluate the effect of Hopkins loss on speech, text, and image data in two scenarios: classification and dimensionality reduction using AEs.

The paper is organized as follows. First, Section \ref{sec_related_work} gives an overview of literature relevant to the present study, namely $H$ (Section \ref{subsec_hopkins_statistic}) and ML methods related to feature space topology (Section \ref{subsec_feature_space_topology_methods}). Then, Hopkins loss is detailed in Section \ref{sec_hopkins_loss}, after which experimental details are given in Section \ref{sec_experiments}. Finally, results are presented in Section \ref{sec_results}, and conclusions are drawn in Section \ref{sec_conclusion}.

\section{Related Work} \label{sec_related_work}

\subsection{Hopkins statistic} \label{subsec_hopkins_statistic}

$H$ is a statistical hypothesis test measuring the clustering tendency of a dataset \cite{hopkins_original}. In $H$, the null hypothesis is that the data is generated by a Poisson point process (uniformly randomly distributed), while the alternative hypothesis is that the data points are clustered \cite{hopkins_original, hopkins_chemical_problems}. Values of $H$ are in the range $[0,1]$, where values greater than 0.75 indicate a clustering tendency at a 90\% confidence level \cite{hopkins_chemical_problems}. Hopkins and Skellam \cite{hopkins_original} introduced $H$ to study the 2D spatial distribution (i.e. coordinates) of trees within a given area. Cross and Jain \cite{hopkins_cross_jain} studied the use of $H$ as a pre-processing step before subjecting data to a clustering algorithm, experimenting with low-dimensional real and simulated datasets. Lawson and Jurs \cite{hopkins_chemical_problems} extended $H$ for both artificial and real chemical datasets. Banerjee and Davé \cite{validating_clusters_hopkins_statistic} used $H$ to validate the presence of meaningful clusters in data using simulated 2D datasets.

$H$ has also been used in more recent studies. In a study by Li et al. \cite{awareness_line_of_sight_hopkins_statistic}, the authors developed a method to enhance indoor localization accuracy by identifying direct and indirect signal conditions using $H$. They used $H$ to measure the spatial randomness of the received signal strength data. Zhang et al. \cite{density_peaks_clustering_hopkins_statistic} proposed a novel density peaks clustering algorithm that integrates $H$ to address the limitations of other similar clustering algorithms related to density measurement and cluster center identification. In their study, $H$ was used to determine whether the data points were randomly distributed or clustered in order to identify potential cluster centers.

\subsection{Machine learning related to feature space topology} \label{subsec_feature_space_topology_methods}

The concept of feature space topology has appeared in ML literature multiple times. Hu et al. \cite{topoloss} proposed a loss function for image segmentation that aims to preserve the feature topology of the reference segmentation, demonstrating effectiveness across various natural and biomedical datasets. Brüel-Gabrielsson et al. \cite{topology_layer} introduced a differentiable topology layer that uses persistent homology to first analyze the input data at multiple scales, and then preserve its topology. They demonstrated the usefulness of their method for tasks where the shape and structure of the data are important, such as incorporating topological priors in generative models. The topological loss function presented by Malyugina et al. \cite{topological_loss_function_image_denoising} is also based on persistent homology, and it was designed for image denoising in low-light conditions. They combined their loss function with L1 and L2 losses to ensure that the denoised image maintains the same topology as the original noisy image, and they showed that their approach helped to reduce artifacts and preserve important structural elements of the input image.

Alberti et al. \cite{manifold_learning_vae} proposed a method for representing manifolds of arbitrary topology using a mixture model of variational AEs. Their method learns the manifold's topology from the data and solves inverse problems by minimizing a data fidelity term restricted to the manifold, demonstrating effectiveness in applications such as deblurring and electrical impedance tomography. Mi et al. \cite{topology_preserving_adversarial} proposed an adversarial training approach based on a topology-preserving regularization term that maintains neighborhood relationships in the representation space, demonstrating enhanced classification accuracy on multiple image datasets.

\section{Hopkins Loss} \label{sec_hopkins_loss}

\begin{figure}[t]
  \centering
  \includegraphics[width=0.49\textwidth]{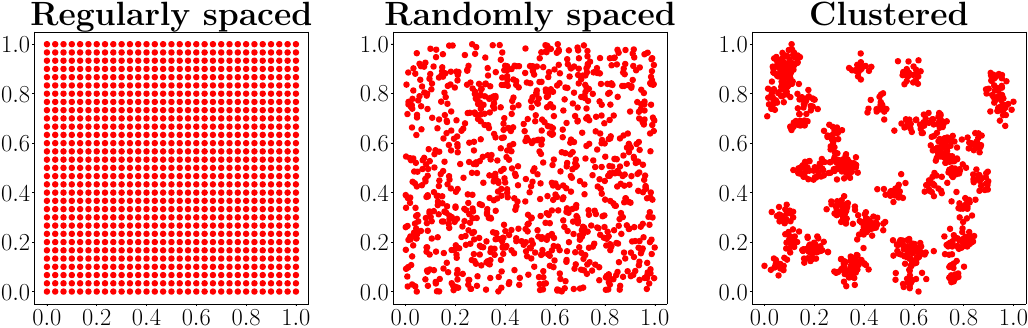}
  \caption{An example 2D visualization of regularly-spaced (left, $H \in [0.01,0.3]$), randomly-spaced (middle, $H \approx 0.5$), and clustered data (right, $H \in [0.7,0.99]$) using the distance metric $D = \text{Chebyshev}$.}
  \label{fig:hopkins_statistic_example_visualization}
\end{figure}

$H$ is computed as follows \cite{hopkins_chemical_problems, validating_clusters_hopkins_statistic}:
\begin{enumerate}
    \item Let $X$ be a set of $n$ $d$-dimensional data points.
    \item Select a random sample of $m \ll n$ data points sampled from $X$ without replacement, denoted as $\tilde{X}$.
    \item Generate a set $Y$ of $m$ data points uniformly distributed at random.
    \item Define two distance measures, both computed using distance metric $D$:
    \begin{enumerate}
        \item $u_i$: The minimum distance of $y_i \in Y$ from its nearest neighbor in $X$.
        \item $w_i$: The minimum distance of $\tilde{x}_i \in \tilde{X}$ from its nearest neighbor in $X$.
    \end{enumerate} 
    \item Calculate the statistic as: $ H = \frac{\displaystyle\sum_{i=1}^{m} u_i}{\displaystyle\sum_{i=1}^{m} u_i + \displaystyle\sum_{i=1}^{m} w_i}$ \hfill (1)
\end{enumerate}
The computation of $H$ leads to values of $H \approx 0.5$ for randomly-spaced data, $H \in [0.01,0.3]$ for regularly-spaced data, and $H \in [0.7,0.99]$ for clustered data \cite{validating_clusters_hopkins_statistic}. Figure \ref{fig:hopkins_statistic_example_visualization} shows examples of data with different values of $H$. In our preliminary experiments using simulated data with different feature topologies, $n \in [1k,100k]$, and $d \in [2,512]$, out of the tested $D \in \{ \text{Euclidean}, \text{Manhattan}, \text{cosine}, \text{Mahalanobis}, \text{Chebyshev} \}$ only Chebyshev distance was able to maintain the desired\footnote{$H \in [0.01,0.3]$ for regularly-spaced data etc.} properties of $H$ across all tested conditions. Consequently, $D = \text{Chebyshev}$ in the present study.

By calculating $H$ using differentiable operations, $H$ can be integrated into a loss function. Consequently, we define the Hopkins loss, $L_\text{H}$, as
\begin{equation} \label{eqn_hopkins_loss}
    L_\text{H} = \left| H - H_\text{T} \right| \ ,
\end{equation}
where $H_\text{T}$ is a pre-defined target value for $H$. Minimizing $L_\text{H}$ causes the neural network-based model to transform the feature topology of the input features towards the topology defined by $H_\text{T}$. For example, with $H_\text{T} = 0.01$, minimizing $L_\text{H}$ leads to the model to learn regularly-spaced features.

In our preliminary experiments, we observed that generating the randomly-distributed set $Y$ within the feature space of the minibatch samples (i.e., set $X$) led to better and more consistent results. Therefore, the values in $Y$ are scaled and shifted using the minimum and maximum values of $X$ before computing $u_i$ in order to better align with the feature space of $X$. Furthermore, Ripley \cite{modelling_spatial_patterns_m_less_than_n} suggested that $m < 0.1n$, and Dubes and Jain \cite{hopkins_approximate_beta_distribution} suggested using $m = 0.05n$ to ensure that nearest-neighbor distances are independent, thereby approximating a Beta distribution. We did not find any difference in $m=kn$ between $k \in \{ 0.02, 0.05, 0.08 \}$ in our preliminary experiments, which led to the selection of $m=0.05n$ for the present experiments, following Dubes and Jain \cite{hopkins_approximate_beta_distribution}.

\section{Experiments} \label{sec_experiments}

We experimented with $L_\text{H}$ using speech, text, and image data (Section \ref{subsec_data}) in both classification (Section \ref{subsec_classification_experiments}) and dimensionality reduction using AEs (Section \ref{subsec_ae_experiments}). The aim was to determine whether the use of $L_\text{H}$ improves, degrades, or does not affect classification performance, and to assess the impact of $L_\text{H}$ on feature topology (in terms of $H$). We implemented the code using PyTorch version 1.13.1, and we used an NVIDIA Tesla V100 GPU to train our models. Our implementation is publicly available on GitHub.\footnote{\url{https://github.com/SPEECHCOG/hopkins_loss}}

\subsection{Data} \label{subsec_data}

The Ryerson Audio-Visual Database of Emotional Speech and Song (RAVDESS) \cite{ravdess} is a multimodal dataset comprising 7,356 recordings performed by 24 professional actors. For the purposes of this study, we utilized 1,440 speech-only recordings from RAVDESS, including eight distinct emotional categories. For each recording, eGeMAPS \cite{egemaps} feature vectors with a dimensionality of $d = 88$ were extracted using the openSMILE toolkit \cite{opensmile}, and the feature vectors were z-score normalized at the dataset level. We randomly split the speaker-wise data into training, validation, and test sets in a 60:20:20 ratio, ensuring that no speaker's data appeared in more than one set.

The IMDB movie review dataset \cite{imdb_dataset} is designed for binary text sentiment classification, containing an even number of positive and negative movie reviews for a total of 50,000 highly polar reviews (25,000 for training and 25,000 for testing). The pre-trained BERT \cite{bert} tokenizer and model were used from the Hugging Face Transformers library \cite{huggingface} to first tokenize the text data, and then to convert the CLS tokens into outputs of dimensionality $d = 768$. We randomly split the 25,000 training reviews into training and validation sets in an 80:20 ratio.

The Fashion-MNIST dataset \cite{fashion_mnist} consists of 70,000 grayscale images of 10 distinct fashion products, with 7,000 images per category and each image sized 28x28 pixels. The dataset includes 60,000 training images and 10,000 testing images, and it was normalized at dataset-level to have pixel values in the range $[-1.0,1.0]$. Each image was flattened to obtain feature vectors of dimensionality $d = 784$. The 60,000 training images were randomly split into 50,000 training and 10,000 validation samples.

\subsection{Classification experiments} \label{subsec_classification_experiments}

We trained simple multi-layer perceptron (MLP) networks for speech emotion recognition (SER) on RAVDESS, text sentiment classification (TSC) on the IMDB movie review dataset, and image classification (IC) on Fashion-MNIST. The MLP network consisted of two fully-connected Gaussian error linear unit (GELU) \cite{gelu_original} layers followed by a linear layer and a softmax function. The GELU layers were followed by a dropout of 20\%. The model output dimensionalities were 128, 128, and $C$, where $C \in \{ 2, 8, 10 \}$ represents the number of classes for text, speech, and image data, respectively.

For model training, we used a combination of two loss functions, categorical cross-entropy loss, $L_\text{CE}$, and $L_\text{H}$:
\begin{equation} \label{eqn_classification_loss}
    L = w_\text{C}L_\text{CE} + (1 - w_\text{C})L_\text{H} \ .
\end{equation}
Here, $w_\text{C} \in [0, 1]$ is the weight for the classification loss $L_\text{CE}$. For example, with $w_\text{C}=1$, $L_\text{H}$ is not utilized during model training, whereas with $w_\text{C}=0.5$ the terms $L_\text{CE}$ and $L_\text{H}$ in Equation \ref{eqn_classification_loss} are given equal weight during model training. Based on preliminary experiments, we found it most effective to compute $L_\text{H}$ from the output of the second MLP layer, and to use $w_\text{C}=0.75$ when both $L_\text{CE}$ and $L_\text{H}$ were used simultaneously.

We used a minibatch size of 1024 samples and an initial learning rate of $10^{-4}$ with a reduction factor of 0.5 based on the validation loss with a patience of 30 epochs. Adam \cite{adam_original} optimizer and early stopping with a patience of 100 epochs based on the validation loss were used, and the model with the lowest validation loss was then used for testing model performance on the test set. Since the class categories in all three datasets of the present experiments had balanced class distributions, accuracy was used as the performance metric on all three tasks. Using all three datasets, we tested all possible combinations of $H_\text{T} \in \{ 0.01, 0.5, 0.99 \}$ and $w_\text{C}=0.75$. As a baseline, we used $w_\text{C}=1.0$. Each experiment was repeated 100 times to account for variability in random model initialization.

\begin{figure}[t]
  \centering
  \includegraphics[width=0.33\textwidth]{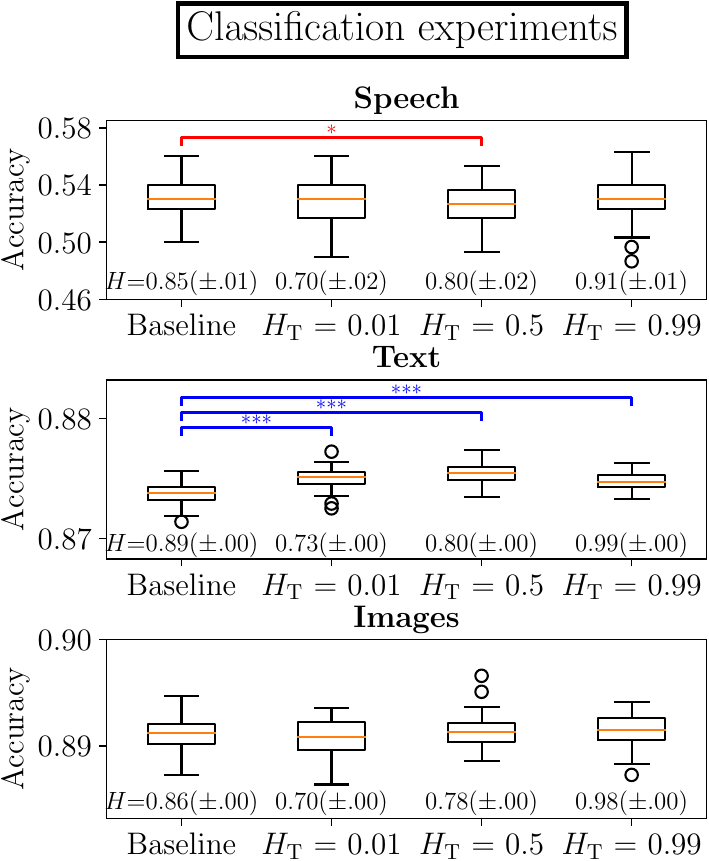}
  \caption{The results of the classification experiments. The Hopkins loss target values $H_\text{T} = 0.01$, $H_\text{T} = 0.5$, and $H_\text{T} = 0.99$ aim at learning regularly-spaced, randomly-spaced, and clustered features, respectively. Statistically significant differences are reported using the Mann-Whitney \textit{U} test \cite{mann_whitney_u_test_original} with either $*$ ($p < 0.05$), $**$ ($p < 0.01$), or $***$ ($p < 0.001$). Results significantly higher than the baseline are marked with \textcolor{blue}{blue}, and vice versa with \textcolor{red}{red}. The number below each box plot is the mean $H$ value ($\pm$95\% confidence interval).}
  \label{fig:hopkins_loss_classification_experiment_results}
\end{figure}

\begin{figure*}[t]
  \centering
  \includegraphics[width=0.99\textwidth]{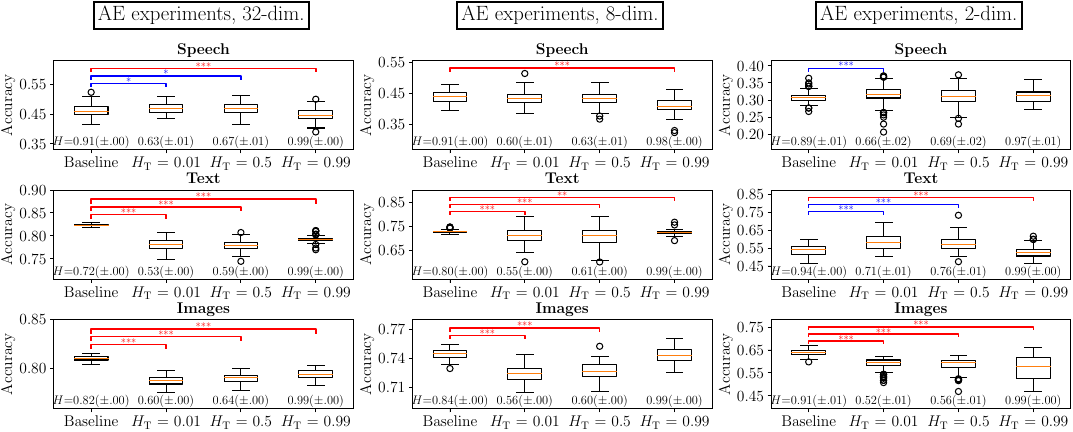}
  \caption{The results of the AE experiments (left: bottleneck feature dimensionality $B=32$, middle: $B=8$, right: $B=2$). The Hopkins loss target values $H_\text{T} = 0.01$, $H_\text{T} = 0.5$, and $H_\text{T} = 0.99$ aim at learning regularly-spaced, randomly-spaced, and clustered features, respectively. Statistically significant differences are reported using the Mann-Whitney \textit{U} test \cite{mann_whitney_u_test_original} with either $*$ ($p < 0.05$), $**$ ($p < 0.01$), or $***$ ($p < 0.001$). Results significantly higher than the baseline are marked with \textcolor{blue}{blue}, and vice versa with \textcolor{red}{red}. The number below each box plot is the mean $H$ value ($\pm$95\% confidence interval).}
  \label{fig:hopkins_loss_ae_experiment_results}
\end{figure*}

\subsection{Autoencoder experiments} \label{subsec_ae_experiments}

\begin{figure}[t]
  \centering
  \includegraphics[width=0.49\textwidth]{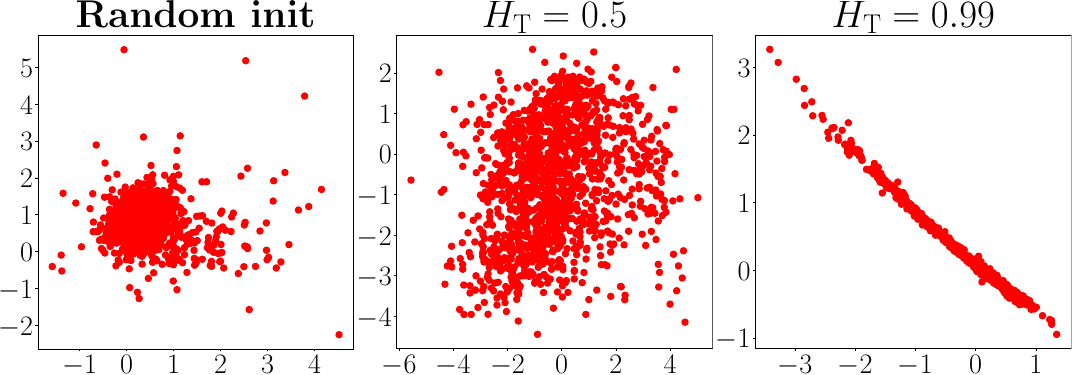}
  \caption{Example of a 0.10 difference in $H$ value when bottleneck feature dimensionality $B = 2$ for the RAVDESS dataset. The bottleneck features of a randomly initialized model (left, $H=0.89$), a trained model with $L_\text{H}$ and $H_\text{T}=0.5$ (middle, $H=0.79$), and a trained model with $L_\text{H}$ and $H_\text{T}=0.99$ (right, $H=0.99$) are shown.}
  \label{fig:2D_AE_speech_visualizations}
\end{figure}

We trained AEs to first compress the input features' dimensionality for each of the three datasets, after which we extracted bottleneck features of the entire dataset using the trained AE. Then, to assess feature degradation, we tested the classification performance of the bottleneck features using a linear classifier for SER on RAVDESS, TSC on the IMDB movie review dataset, and IC on Fashion-MNIST. The AE consisted of an encoder and a decoder, and the encoder was an MLP network consisting of two fully-connected GELU layers followed by a linear layer, and the first two layers were followed by a 20\% dropout. The encoder output dimensionalities were 128, 128, and $B$, where $B \in \{ 32, 8, 2 \}$ stands for the bottleneck feature dimensionality. The decoder was simply a mirrored version of the encoder.

For AE training, we used a combination of two loss functions, $L_\text{H}$ and mean squared error loss, $L_\text{MSE}$, in the following manner:
\begin{equation} \label{eqn_ae_loss}
    L = w_\text{R}L_\text{MSE} + (1 - w_\text{R})L_\text{H} \ .
\end{equation}
Here, $w_\text{R} \in [0, 1]$ is the weight for the reconstruction loss $L_\text{MSE}$. Based on preliminary experiments, we compute $L_\text{H}$ from the output of the bottleneck layer of the encoder, and use $w_\text{R}=0.75$ when both $L_\text{MSE}$ and $L_\text{H}$ were used simultaneously.

After extracting the bottleneck features using the trained AE, an MLP classifier consisting of a single linear layer and a softmax function was trained for classification using the bottleneck features as the model input. For both the AE and classifier training, we used the same hyperparameters and training pipeline as described in Section \ref{subsec_classification_experiments} with the only exceptions of having $4 \cdot 10^{-4}$ as the initial learning rate for both AE and classifier training, and without including $L_\text{H}$ in classifier training. We tested all possible combinations of $H_\text{T} \in \{ 0.01, 0.5, 0.99 \}$, $B \in \{ 32, 8, 2 \}$, and $w_\text{R}=0.75$. As a baseline, we tested all combinations of $B \in \{ 32, 8, 2 \}$ and $w_\text{R}=1.0$. Each experiment was repeated 100 times to account for the minor variability caused by the random initialization of model weights.

\section{Results} \label{sec_results}

Figure \ref{fig:hopkins_loss_classification_experiment_results} presents the classification experiment results (Section \ref{subsec_classification_experiments}). For speech and image data, integrating $L_\text{H}$ into model training generally does not degrade model performance, while also adding the benefit of modifying the features towards a desired topology. Additionally, for text data, adding $L_\text{H}$ outperformed the baseline for all targeted feature topologies. On average, including $L_\text{H}$ in the classification experiments modified $H$ towards the target value $H_\text{T}$ by approx. 0.09, 0.11, and 0.12 in terms of $H$ value for speech, text, and image data, respectively.

Figure \ref{fig:hopkins_loss_ae_experiment_results} shows the AE experiment results (Section \ref{subsec_ae_experiments}). In most cases, the use of $L_\text{H}$ resulted in lower classification performance compared to the baseline, which is natural since $L_\text{H}$ was not part of the classifier training in AE experiments. However, there were multiple cases where the use of $L_\text{H}$ resulted in either a similar or even higher classification performance than the baseline. In AE experiments, the modification in feature space topology was notably larger than in the classification experiments: On average, the inclusion of $L_\text{H}$ in the AE training modified $H$ towards the target value $H_\text{T}$ by approx. 0.19, 0.18, and 0.22 in terms of $H$ value for speech, text, and image data, respectively. As a reference, Figure \ref{fig:2D_AE_speech_visualizations} illustrates a 0.10 difference in terms of $H$ value in 2D space when using $L_\text{H}$ with AEs. Note that since the samples are already organized into a single large cluster for the randomly-initialized model, with $L_\text{H}$ and $H_\text{T}=0.99$ it is easier to optimize forming one large cluster than multiple clusters during model training. Overall, the average decrease in performance in the AE experiments is relatively small compared to the average change in $H$ values.

Interestingly, for both classification (Figure \ref{fig:hopkins_loss_classification_experiment_results}) and AE (Figure \ref{fig:hopkins_loss_ae_experiment_results}) experiments, the feature topologies for all baseline models were clustered in terms of $H$ value, indicating a natural bias towards a clustered topology in the original features. This may explain why $L_\text{H}$ could not force the features into a regularly-spaced topology ($H \in [0.01,0.3]$) with $H_\text{T}=0.01$. Another potential explanation for why $L_\text{H}$ could not enforce a regularly-spaced topology is that this may be inherently a more difficult task for the model: structuring features uniformly might require more effort than modifying them to be randomly spaced or clustered. However, the experiments showed that $L_\text{H}$ can be used to adjust the original clustered topology towards being more regularly spaced.

Table \ref{table:baseline_vs_hopkins_loss_training_time_per_epoch} compares the duration of training epochs with and without the use of $L_\text{H}$. Overall, the computational overhead introduced by $L_\text{H}$ is modest, with an average increase in epoch duration across all three datasets of approximately 10--13\% in all experimental conditions. For the smallest dataset, RAVDESS, the overhead was marginal. In contrast, the two other datasets, each notably larger than RAVDESS in terms of sample count and feature dimensionality, exhibited a more noticeable increase in training time when $L_\text{H}$ was applied. These results suggest that the computational cost of $L_\text{H}$ scales with both dataset size and feature dimensionality. Furthermore, in several cases, the use of $L_\text{H}$ led to a slight increase in the variability of epoch durations, as reflected by a higher standard deviation. However, it is important to note that these results may vary depending on factors such as model complexity, dataset characteristics, and the hardware used for training.

\section{Discussion and Conclusion} \label{sec_conclusion}

\begin{table}[t]
    \centering
    \caption{Comparison of training epoch durations with the use of $L_\text{H}$ and without it (baseline) across classification and AE experiments. No notable differences in epoch durations were observed across different $H_\text{T}$ values, and therefore both the mean and standard deviation (SD) of epoch durations are computed across all training epochs and $H_\text{T}$ configurations. For each experimental condition, the average computational overhead of using $L_\text{H}$ across all three datasets is highlighted in \textcolor{red}{red text}.}
    \includegraphics[width=0.48\textwidth]{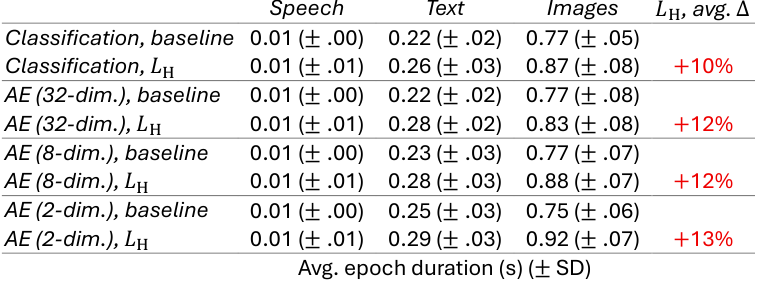}
    \label{table:baseline_vs_hopkins_loss_training_time_per_epoch}
\end{table}

This paper presented a novel loss function, $L_\text{H}$, to modify features into a desired topology. We evaluated $L_\text{H}$ on speech, text, and image data in two scenarios: classification and dimensionality reduction using AEs. As a result, integrating $L_\text{H}$ into classification maintained model performance and, in some cases, even improved it, while also modifying the feature space towards a desired topology. With AEs, the use of $L_\text{H}$ slightly reduced classification performance but notably modified the feature topology. This can be useful in dimensionality reduction applications where feature topology is more important than a slight degradation in classification performance.

These findings suggest that $L_\text{H}$ can be useful in scenarios where controlling feature space topology is beneficial. In particular, our experiments demonstrate its applicability in \textit{classification} and \textit{dimensionality reduction} tasks, where modifying the topology may aid in improving generalization or enhancing the structure of low-dimensional embeddings for e.g. visualization or data compression. Beyond these tested scenarios, $L_\text{H}$ may also hold potential in other applications discussed in Section \ref{sec_introduction}, such as \textit{generative modeling}, \textit{bioinformatics}, \textit{transfer learning}, and \textit{adversarial robustness}.

However, as our experiments were conducted using simple MLPs, the effects of $L_\text{H}$ on classification performance are not fully explored and may vary with more complex models. Potential future work includes applying $L_\text{H}$ to all neural network layers simultaneously, testing additional distance metrics, and exploring further use cases beyond classification and AEs (e.g. those mentioned in Section \ref{sec_introduction}). Lastly, it is important to note that the choice of an optimal distance metric in $L_\text{H}$ can vary notably depending on the selected features and the targeted application. For example, if data is regularly spaced on the surface of a hypersphere, it is regularly spaced in terms of cosine distance but not in terms of Chebyshev distance.

\section*{Acknowledgment}

This work was supported in part by the Research Council of Finland (grant no. 343498) and in part by the Sigrid Jusélius Foundation. The authors would like to thank Professor Okko Räsänen for the fruitful discussions and helpful comments, and Tampere Center for Scientific Computing for the computational resources used in this study. The author Einari Vaaras would also like to thank the Finnish Foundation for Technology Promotion for the encouragement grants and the Nokia Foundation for the Nokia Scholarship.

\bibliographystyle{IEEEtran}
\bibliography{mybib.bib}

\end{document}